\documentclass{article}

 \usepackage[dblblindworkshop, final, nonatbib]{neurips_2025}
\workshoptitle{The Second Workshop on GenAI for Health: Potential, Trust, and Policy Compliance}

\usepackage[utf8]{inputenc} 
\usepackage[T1]{fontenc}    
\usepackage{hyperref}       
\usepackage{url}            
\usepackage{booktabs}       
\usepackage{amsfonts}       
\usepackage{nicefrac}       
\usepackage{microtype}      
\usepackage{xcolor}         
\usepackage[most]{tcolorbox}

\newtcolorbox{promptbox}[2][]{
  breakable,
  title=#2,
  width=\linewidth,
  colback=gray!5,
  colframe=gray!50!black,
  boxrule=0.8pt,
  arc=4pt,
  left=4pt, right=4pt, top=4pt, bottom=4pt,
  fontupper=\small\ttfamily,
  #1
}

\newtcolorbox{infobox}[2][]{
  breakable,
  title=#2,
  width=\linewidth,
  colback=gray!5,
  colframe=gray!50!black,
  boxrule=0.8pt,
  arc=4pt,
  left=4pt, right=4pt, top=4pt, bottom=4pt,
  #1
}

\usepackage{siunitx}  
\usepackage{amsmath}
\usepackage{graphicx}
\usepackage{cleveref}
\usepackage{multirow}

\usepackage{pgfmath}
\usepackage[table]{xcolor}
\newcommand{\RankColor}[2]{
    \pgfmathtruncatemacro{\percent}{((((22-#1)*(22-#1)-1)/(22*22-1))*100+5)*0.7}%
    \edef\colorcmd{\noexpand\cellcolor{blue!\percent}}%
    \colorcmd #2%
}

\usepackage{adjustbox} 
\usepackage[
    backend=biber,
    style=numeric,
    sorting=none,
    maxbibnames=4
]{biblatex}
\usepackage{booktabs}   
\usepackage{makecell}   

\title{H-DDx: A Hierarchical Evaluation Framework \\ for Differential Diagnosis}

\author{
    \textbf{Seungseop Lim$^{1}$\thanks{\ Equal contribution.}, Gibaeg Kim$^{1}$\footnotemark[1], Hyunkyung Lee$^{1}$}, \\
    \textbf{Wooseok Han$^{1}$, Jean Seo$^{1}$, Jaehyo Yoo$^{1}$, Eunho Yang$^{1,2}$}\thanks{\ Corresponding author.} \\
    $^{1}$AITRICS \qquad
    $^{2}$KAIST \\
    \texttt{\{ss.lim, gb.kim, eunhoy\}@aitrics.com, eunhoy@kaist.ac.kr}
}

\begin{document}

\maketitle

\begin{abstract}
An accurate differential diagnosis (DDx) is essential for patient care, shaping therapeutic decisions and influencing outcomes. Recently, Large Language Models (LLMs) have emerged as promising tools to support this process by generating a DDx list from patient narratives. However, existing evaluations of LLMs in this domain primarily rely on flat metrics, such as Top-k accuracy, which fail to distinguish between clinically relevant near-misses and diagnostically distant errors. To mitigate this limitation, we introduce \textbf{H-DDx}, a hierarchical evaluation framework that better reflects clinical relevance. H-DDx leverages a retrieval and reranking pipeline to map free-text diagnoses to ICD-10 codes and applies a hierarchical metric that credits predictions closely related to the ground-truth diagnosis. In benchmarking 22 leading models, we show that conventional flat metrics underestimate performance by overlooking clinically meaningful outputs, with our results highlighting the strengths of domain-specialized open-source models. Furthermore, our framework enhances interpretability by revealing hierarchical error patterns, demonstrating that LLMs often correctly identify the broader clinical context even when the precise diagnosis is missed.
\end{abstract}

\section{Introduction}
Differential diagnosis (DDx) is a fundamental component of clinical reasoning, involving the systematic consideration of multiple possible conditions that could explain a patient’s symptoms. This process is central to comprehensive case evaluation, as it enables the detection of critical yet subtle conditions, guides diagnostic testing, and promotes the efficient use of resources. Moreover, DDx enhances communication, fosters patient trust, and mitigates cognitive biases by compelling clinicians to consider a broad set of plausible explanations \cite{tierney2007, guyatt2002, rhoads2022}. Despite its importance, medical diagnosis remains a cognitively demanding and high-stakes task, particularly in its early stages when uncertainty is greatest. Recently, Large Language Models (LLMs) have emerged as promising tools to support this process by generating a DDx list from patient narratives \cite{thirunavukarasu2023large, nashwan2023harnessing}. 

To assess how effectively LLMs can assist clinicians in the DDx process, rigorous evaluation is essential. However, existing evaluations of LLMs for DDx rely predominantly on simplistic metrics that inadequately reflect clinical utility, with most prior works~\cite{mcduff2025towards, chen2025enhancing, rios2024evaluation, bhasuran2025preliminary} focusing primarily on Top-k accuracy, which merely checks whether the final diagnosis appears in the DDx set. These approaches face two fundamental limitations. First, they reduce evaluation to mere inclusion or surface-level overlap, failing to capture the clinical value of the DDx set as a whole. A list that contains the ground-truth diagnosis but is cluttered with irrelevant or misleading suggestions may erode clinician trust and provide little practical support. 
Second, when computing Top-k accuracy, the determination of whether a predicted disease matches the ground-truth is often delegated to an LLM judge, introducing ambiguity and a lack of transparency. Because the matching criteria are implicit and prone to variation across different LLM judges or prompts, assessments become subjective and difficult to reproduce. 
This black-box paradigm fails to differentiate the severity of errors, lacking a principled or hierarchical framework. For instance, suggesting a viral upper respiratory infection for a case of influenza constitutes a clinically minor error, as both are respiratory conditions with overlapping symptoms. In contrast, suggesting a neurological condition like migraine for the same influenza case represents a far more significant diagnostic error, indicating a fundamental misunderstanding of the clinical presentation. Without such a structure, evaluations struggle to distinguish consistently and interpretably between clinically distinct errors.

\begin{figure}[t]
  \centering
  \includegraphics[width=0.9\linewidth]{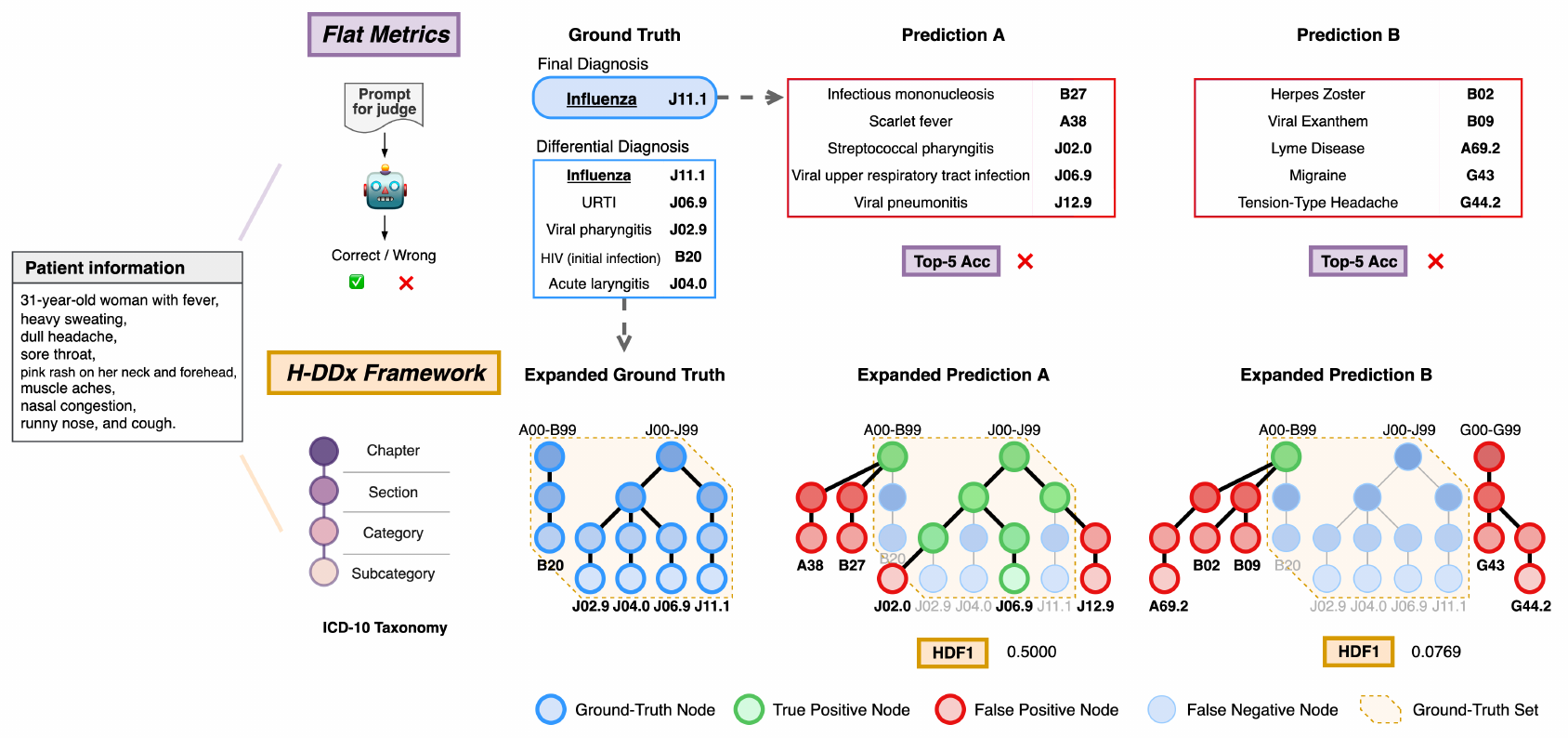}
    \caption{
    Comparison of the H-DDx framework and conventional flat metrics. For a patient with Influenza, flat metrics fail to distinguish between the clinically relevant DDx set from Prediction A (related respiratory infections) and the irrelevant list from Prediction B (e.g., Migraine), scoring both poorly. H-DDx uses the ICD-10 taxonomy for a more nuanced evaluation. By expanding differential diagnosis sets into the taxonomy, it identifies Prediction A's outputs as clinically relevant near-misses, while Prediction B's are distant errors. HDF1 score quantifies this distinction, capturing the superior clinical utility of Prediction A that flat metrics overlook.
    } 
  \label{fig:fig1}
\end{figure}

To address these limitations, we propose \textbf{H-DDx}, a novel hierarchical evaluation framework for a more clinically meaningful assessment of differential diagnosis. As illustrated in Figure~\ref{fig:fig1}, conventional flat metrics fail to distinguish between clinically relevant near-misses and distant diagnostic errors, treating all incorrect predictions equally. This is a critical gap our framework aims to fill. By leveraging the structured knowledge within the International Classification of Diseases 10th Revision (ICD-10) taxonomy, H-DDx moves beyond simple accuracy to reward these diagnostically relevant suggestions. This allows for a more nuanced and interpretable analysis of an LLM's true clinical reasoning capabilities, revealing insights that conventional metrics overlook.

Through large-scale evaluation of 22 leading LLMs on a public benchmark, DDXPlus~\cite{fansi2022ddxplus}, we demonstrate the framework's effectiveness in providing a deeper understanding of model performance. Our main contributions are summarized as follows:

\begin{itemize}
    \item We introduce \textbf{H-DDx}, a novel hierarchical evaluation framework that maps free-text diagnoses to the ICD-10 taxonomy and applies a hierarchical metric for a more clinically nuanced assessment.
    \item Through a large-scale evaluation of 22 leading LLMs, we demonstrate that conventional flat metrics misrepresent models' differential diagnostic capabilities by penalizing clinically plausible, near-miss diagnoses as severely as entirely incorrect ones, thereby undervaluing domain-specialized models.
    \item Our framework enhances the interpretability of model behavior by identifying and analyzing hierarchical patterns in diagnostic performance, revealing that LLMs often grasp the correct clinical context even when they fail to pinpoint the exact diagnosis.
\end{itemize}

\section{Related Work}
\subsection{LLMs for Differential Diagnosis}
The application of Large Language Models (LLMs) to augment clinical decision-making has recently gained significant traction, particularly in the domain of differential diagnosis (DDx)~\cite{thirunavukarasu2023large, nashwan2023harnessing, dinc2025comparative}. Numerous studies have highlighted the potential of LLMs to generate a comprehensive DDx list from patient narratives and clinical vignettes~\cite{mcduff2025towards, zhou2025large}. The evaluation of these models has largely relied on conventional flat metrics such as Top-k accuracy, assessing whether the final diagnosis appears within the top predictions. While useful, this approach fails to credit clinically relevant near-misses, treating all diagnostic errors as equally severe. Our work moves beyond this limitation by introducing a hierarchical evaluation framework that captures the nuances of clinical reasoning.

\subsection{Automated ICD Coding}
Automated assignment of International Classification of Diseases (ICD) codes to clinical documentation is a well-established task in medical natural language processing~\cite{mullenbach2018explainable}. Early approaches often utilized rule-based systems or traditional machine learning models, while more recent work has demonstrated the superior performance of deep learning architectures, including LLMs~\cite{shi2017towards, carberry2024gpt, li2018automated}. Our work is distinct from automated coding; rather than predicting codes for billing or administrative purposes, we focus on developing a high-fidelity mapping from LLM-generated free-text diagnoses to a standardized taxonomy. This mapping serves as a foundational step to enable a more sophisticated, semantically aware evaluation of the models' diagnostic capabilities.

\subsection{Hierarchical Classification}
Hierarchical classification, where class labels are organized in a taxonomy, is a long-standing area of machine learning research~\cite{silla2011survey}. Recognizing that not all errors are equal, researchers have developed specialized metrics that account for the hierarchical structure, such as hierarchical precision and recall~\cite{kosmopoulos2015evaluation, riehl2023hierarchical, lanus2025hierarchical, riehl2023hierarchical}. These metrics assign partial credit for misclassifications that are taxonomically close to the true label, providing a more fine-grained performance assessment than conventional flat metrics. Our framework leverages these established hierarchical evaluation principles, adapting them specifically to the clinical context of differential diagnosis assessment, where both the multi-label nature and the semantic relationships between diagnoses must be considered.

\begin{figure}[b]
  \centering
  \includegraphics[width=1.0\linewidth]{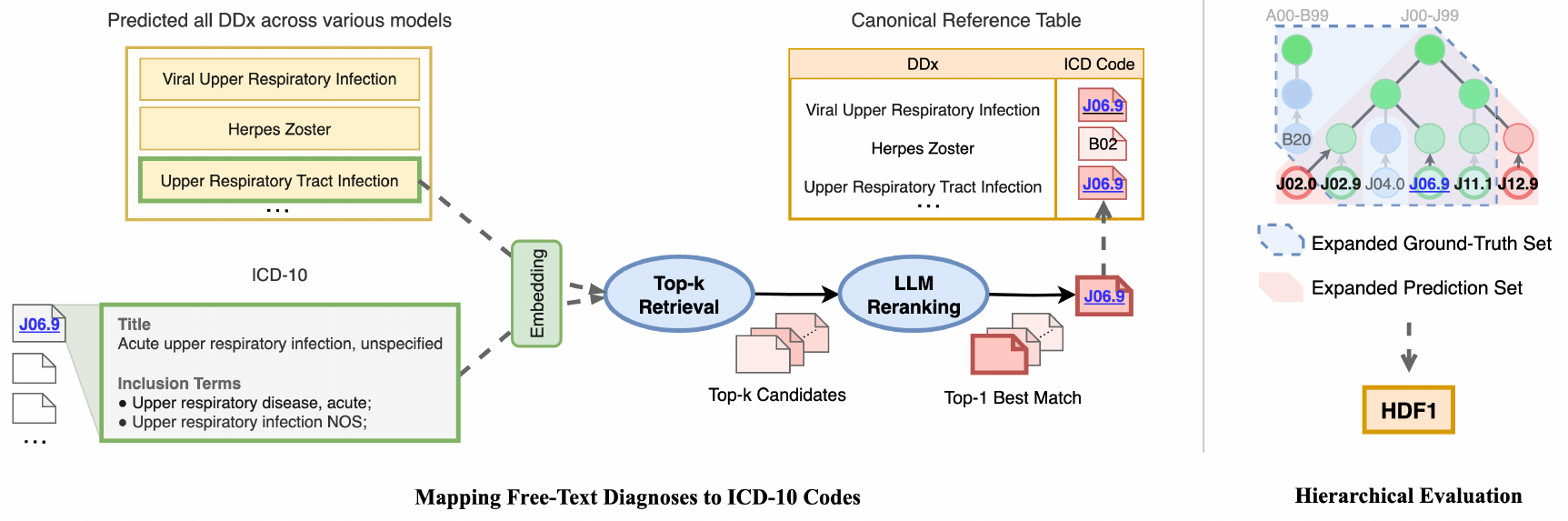} 
  \caption{Overview of the H-DDx framework.}
  \label{fig:fig2}
\end{figure}

\section{The H-DDx Framework}
This section presents \textbf{H-DDx}, a hierarchical framework for evaluating the DDx capabilities of LLMs. Our framework consists of two steps: (i) a mapping pipeline that combines embedding-based retrieval with LLM reranking for high-accuracy conversion from free-text diagnoses to the ICD-10 codes, and (ii) the Hierarchical DDx F1 (HDF1) metric that performs set-based comparisons using ancestral code expansion within the ICD-10 taxonomy. The overall framework is illustrated in Figure~\ref{fig:fig2}. 

\subsection{Leveraging the ICD-10 Taxonomy}
To quantify the clinical distance between predicted and ground-truth differential diagnoses, our H-DDx framework leverages the inherent hierarchy of the ICD-10. This globally adopted taxonomy structures medical conditions into a four-level tree, providing a standardized basis for our evaluation:

\begin{itemize}
    \item \textbf{Chapter:} Groups diseases by major category, such as anatomical system (e.g., J00-J99: Diseases of the Respiratory System).
    \item \textbf{Section:} Clusters related conditions within a chapter (e.g., J09-J18: Influenza and Pneumonia).
    \item \textbf{Category:} Specifies distinct diseases with a three-character code (e.g., J11: Influenza due to unidentified influenza virus).
    \item \textbf{Subcategory:} Offers the highest granularity for detailed classification (e.g., J11.1: Influenza due to unidentified influenza virus with other respiratory manifestations).
\end{itemize}
By grounding our evaluation in this structure, H-DDx can measure not just whether a diagnosis is correct, but how close an incorrect diagnosis is to the correct one. 
This hierarchical organization is not arbitrary but reflects clinical relationships. Conditions within the same branch typically share anatomical systems, pathophysiological mechanisms, or etiological origins. This medical grounding makes the ICD-10 taxonomy an ideal foundation for assessing the clinical quality of DDx predictions beyond binary correct/incorrect judgments.

\subsection{Mapping Free-Text Diagnoses to ICD-10 Codes} 
\label{ssec:mapping}

Hierarchical evaluation requires mapping free-text diagnosis from LLMs to a standardized clinical taxonomy, such as ICD-10. Our mapping pipeline combines embedding-based retrieval and LLM-based reranking for high accuracy. For each free-text diagnosis, we first use the \texttt{text-embedding-3-large} model to retrieve the top-k candidate ICD-10 codes. An LLM-based reranker then identifies the best match from these candidates. We validated this approach on a benchmark set of 101 diagnoses, manually mapped by clinicians, achieving a Top-1 accuracy of 93.1\%. The reranking stage proved critical, improving upon the retrieval-only accuracy of 71.3\%. Detailed performances are provided in Table~\ref{tab:table1}.

To ensure a fair comparison across all models, we first compile a comprehensive set of all unique DDx generated across all models evaluated. This unified set was then mapped to ICD-10 codes to create a single, canonical reference table. By applying this same mapping table to every model's output, we eliminate any potential bias from the mapping process itself.

\begin{table}[b]
\centering
\caption{Comparison of models for mapping free-text diagnoses to ICD-10 codes. The table presents the Top-1 accuracy of various embedding models for retrieval and the subsequent improvements from applying LLM rerankers to the best-performing retriever.}
\resizebox{0.6\linewidth}{!}{

\begin{tabular}{llS[table-format=1.4]}
\toprule
\textbf{ } & \textbf{Model} & {\textbf{Top-1}} \\
\midrule
\multirow{5}{*}{\parbox{2.5cm}{\centering \textbf{Top-k Retrieval}}}
& biobert-v1.1 & 0.4653 \\
& BioSimCSE-BioLinkBERT-BASE & 0.5446 \\
& Qwen3-Embedding-0.6B & 0.5842 \\
& pubmedbert-base-embeddings & 0.6436 \\
& text-embedding-3-large & 0.7129 \\
\midrule
\multirow{4}{*}{\parbox{2.5cm}{\centering \textbf{LLM Reranking} \\ \textit{(on best retriever)}}}
& + gemini-2.5-flash-lite & 0.8317 \\
& + gpt-4o-mini & 0.8614 \\
& + gemini-2.5-flash & 0.8713 \\
& + gpt-4o & \textbf{0.9307} \\
\bottomrule
\end{tabular}

}
\label{tab:table1}
\end{table}

An error analysis of the pipeline's failures revealed that they were not random errors, but systematic and clinically adjacent near-misses. Since all observed errors were correct at the immediate parent level in the ICD-10 taxonomy, these minor inaccuracies do not compromise the validity of our hierarchical evaluation. This finding confirms the pipeline's robustness for its intended use. Further details are available in \cref{app:mapping-pipeline}. 

\subsection{A Hierarchical Evaluation Metric}
\label{ssec:hdf1}

The core of our framework is the Hierarchical DDx F1 (HDF1) metric, an extension of the standard F1-score designed to measure clinical utility by leveraging the ICD-10 taxonomy. Whereas conventional flat metrics treat diagnoses as isolated labels, HDF1 evaluates them within their ancestral context in the ICD-10 taxonomy. This enables a more nuanced and clinically grounded assessment that reflects clinical realities.

HDF1 builds upon established hierarchical precision and recall measures from multi-label hierarchical classification \cite{kiritchenko2006learning}. To formalize this, the calculation of HDF1 begins by expanding the sets of ground-truth and predicted DDx to include all their ancestral nodes in the hierarchy. For a given patient case $i$, let $D_i$ be the ground-truth DDx set and $\hat{D_i}$ be the predicted set. We define an augmentation function, $\text{Augment}(S)$, which augments the set $S$ by adding all ancestral nodes for each diagnosis in $S$, from its immediate parent up to the chapter level in the ICD-10 taxonomy. The augmented sets for comparison are therefore $C_i = \text{Augment}(D_i)$ and $\hat{C_i} = \text{Augment}(\hat{D_i})$.

Based on these augmented sets, Hierarchical DDx Precision (HDP) and Hierarchical DDx Recall (HDR) are computed. The scores are calculated per patient case and then macro-averaged across the dataset to ensure each case contributes equally. HDF1 is the harmonic mean of the resulting HDP and HDR, as defined in Eqs. \crefrange{eq:hip}{eq:hif}:
\begin{align}
  \mathrm{HDP} &= \frac{1}{N} \sum_{i=1}^{N} \frac{|C_i \cap \hat{C_i}|}{|\hat{C_i}|}\text{,} \label{eq:hip} \\
  \mathrm{HDR} &= \frac{1}{N} \sum_{i=1}^{N} \frac{|C_i \cap \hat{C_i}|}{|C_i|}\text{,}  \label{eq:hir} \\
  \text{and } \mathrm{HDF1} &= \frac{2 \times \mathrm{HDP} \times \mathrm{HDR}}{\mathrm{HDP} + \mathrm{HDR}} \label{eq:hif}
\end{align}
where $N$ is the total number of patient cases.

\section{Experimental Setup}
\subsection{Datasets}
Our evaluation utilizes DDXPlus~\cite{fansi2022ddxplus}, a large-scale synthetic benchmark dataset for differential diagnosis. Recognized for its complex and diverse medical cases, comprising over 1.3 million synthetic Electronic Health Record (EHR) patient records across 49 distinct pathologies. The research community widely adopts it to benchmark models for diagnosis tasks~\cite{chen2023llm, li2024knowledge, tam2024let, xie2024preliminary}. The dataset features comprehensive patient information, including socio-demographics, underlying conditions, symptoms, and antecedents (i.e., medical history). Each patient record contains approximately 10 symptoms and 3 antecedents on average, which helps bridge the symptom data gap prevalent in real-world EHR datasets like MIMIC. To create a cost-effective yet representative test set, we selected 730 cases from the full DDXPlus dataset via stratified sampling~\cite{neyman1992two}. This approach ensures that the distribution of pathologies in our subset faithfully reflects that of the original benchmark, providing a robust foundation for our evaluation.

\subsection{Models}
In this work, we evaluated the performance of 22 LLMs, comprising 11 proprietary and 11 open-source models, focusing on their direct, non-reasoning capabilities.
For proprietary models, we covered available versions of OpenAI's GPT series~\cite{hurst2024gpt}, Google's Gemini~\cite{team2023gemini}, and Anthropic's Claude~\cite{anthropic2025claude4}. The open-source models included leading models such as Google's Gemma~\cite{team2025gemma}, Alibaba's Qwen~\cite{yang2025qwen3}, and Microsoft's Phi~\cite{abdin2024phi3technicalreporthighly}. Additionally, we evaluated models trained for clinical context such as MedGemma~\cite{sellergren2025medgemma} (adapted from Gemma3) and MediPhi~\cite{corbeil2025modular} (adapted from Phi). A detailed list of the specific model versions used is available in \cref{app:reproducibility}.

\subsection{Evaluation Protocol}
For each patient case, models were prompted to predict a list of five differential diagnoses based on the chief complaint and history of present illness. We evaluated these predictions using both a hierarchical evaluation with our proposed H-DDx framework and a conventional flat evaluation to benchmark against prior work. 

\paragraph{Hierarchical Evaluation} Our primary evaluation is based on the HDF1 score, which requires mapping the varied free-text diagnoses to a standardized taxonomy. To illustrate the necessity of this process, we first quantified the scale and diversity of the raw diagnoses generated by the 22 LLMs. For the DDXPlus test set, the models produced 80,300 diagnoses, comprising 4,905 unique free-text strings. This sheer volume of unstructured terms underscores the need for a systematic mapping pipeline to enable a principled, hierarchical comparison. As detailed in Section~\ref{ssec:mapping}, our pipeline standardizes these outputs into 1,507 unique ICD-10 codes. Based on these standardized codes, we calculate the HDF1 score (detailed in Section~\ref{ssec:hdf1}) to provide a nuanced assessment of clinical relevance.

\paragraph{Flat Evaluation} For comparison, we also computed Top-5 Accuracy, a conventional flat metric. Following prior work~\cite{mcduff2025towards}, we use an LLM to judge the correctness of disease names. Top-5 Accuracy assesses whether the predicted DDx set contains the ground-truth diagnosis. We used \texttt{gpt-4o} to judge semantic equivalence between the predicted and ground-truth. The prompt used for this evaluation, adopted from prior work, is detailed below.

\begin{promptbox}[verbatim]{Top-5 Accuracy Evaluation Prompt}
Is our predicted diagnosis correct (y/n)?\\
Predicted diagnosis: [diagnosis], True diagnosis: [label] Answer [y/n].
\end{promptbox}

A prediction was marked as correct if the model output `y'. Full details for reproducibility are provided in \cref{app:reproducibility}.

\begin{table}[t]
\centering
\caption{Main evaluation results. We compare LLMs using Top-5 Accuracy (Top-5) and our proposed Hierarchical DDx F1-score (HDF1). The $\Delta$ Rank indicates the change in rank from Top-5 Accuracy to HDF1. The best score for each metric is in \textbf{bold}, and the second-best is \underline{underlined}.}
\begin{adjustbox}{width=0.5\textwidth,center}

\begin{tabular}{
    l 
    S[table-format=1.4] 
    S[table-format=1.4] 
    c 
}
    \toprule
    \multicolumn{1}{c}{\textbf{Model}} 
    & \textbf{Top-5} 
    & \textbf{HDF1} 
    & $\boldsymbol{\Delta \textbf{Rank}}$ \\
    \midrule
    \multicolumn{4}{c}{\textbf{Proprietary Models}} \\
    \midrule
    Claude-Haiku-3.5      & 0.7630 & 0.3237 & $\downarrow 3$ \\
    Claude-Sonnet-3.7     & \underline{0.8360} & 0.3380 & $\downarrow 6$ \\
    Claude-Sonnet-4       & \textbf{0.8390} & \textbf{0.3673} & $\text{--}$ \\
    Gemini-2.5-Flash-Lite & 0.7890 & 0.3496 & $\uparrow 2$ \\
    Gemini-2.5-Flash      & 0.8320 & 0.3483 & $\downarrow 2$ \\
    GPT-4o-mini           & 0.7240 & 0.3276 & $\uparrow 3$ \\
    GPT-4o                & 0.8040 & 0.3499 & $\uparrow 1$ \\
    GPT-4.1-nano          & 0.7660 & 0.3213 & $\downarrow 9$ \\
    GPT-4.1-mini          & 0.7590 & 0.3232 & $\downarrow 2$ \\
    GPT-4.1               & 0.8010 & 0.3387 & $\downarrow 2$ \\
    GPT-5                 & 0.7830 & 0.3448 & $\uparrow 1$ \\
    \addlinespace
    \midrule
    \multicolumn{4}{c}{\textbf{Open-source Models}} \\
    \midrule
    Phi-3.5-mini          & 0.6550 & 0.3187 & $\uparrow 1$ \\
    Gemma3-4B             & 0.6080 & 0.2891 & $\text{--}$ \\
    Gemma3-12B            & 0.7180 & 0.3075 & $\downarrow 3$ \\
    Gemma3-27B            & 0.7460 & 0.3225 & $\downarrow 2$ \\
    Qwen2.5-72B           & 0.7420 & 0.3299 & $\uparrow 4$ \\
    Qwen3-4B              & 0.6720 & 0.3291 & $\uparrow 6$ \\
    Qwen3-14B             & 0.7750 & 0.3367 & $\text{--}$ \\
    Qwen3-32B             & 0.7630 & 0.3300 & $\uparrow 2$ \\
    Qwen3-235B-A22B       & 0.7770 & 0.3215 & $\downarrow 10$ \\
    \addlinespace
    \midrule
    \multicolumn{4}{c}{\textbf{Medical Fine-tuned Models}} \\
    \midrule
    MedGemma-27B          & 0.7650 & 0.3310 & $\uparrow 1$ \\
    MediPhi               & 0.6660 & \underline{0.3526} & $\uparrow 18$ \\
    \bottomrule
\end{tabular}

\end{adjustbox}
\label{tab:table2}
\end{table}

\section{Results and Analysis}
Our experimental results indicate that H-DDx offers a more effective evaluation of the diagnostic capabilities of various models. The analysis suggests that the proposed HDF1 metric not only identifies hidden strengths of domain-specialized models previously underestimated by conventional flat metrics but also provides interpretable insights into hierarchical diagnostic patterns. Through comprehensive experiments and case studies, our findings suggest that H-DDx produces more clinically meaningful evaluations that better align with practical healthcare needs.

\subsection{Strengths of Domain-Specialized Models Identified by H-DDx}
\label{sec:rq1}
The most significant impact of applying HDF1 was a substantial reordering of model rankings. Our hierarchical approach reveals the true capabilities of clinically specialized models that were systematically undervalued by flat metrics.
As shown in Table~\ref{tab:table2}, this approach leads to substantial changes in model rankings, particularly for domain-specialized models whose strengths were previously underestimated. A notable example is MediPhi, which improves by 18 ranks, rising from 20th in Top-5 Accuracy to 2nd in HDF1. However, this effect was not observed uniformly across all specialized models. MedGemma showed more modest gains, highlighting the variable impact of domain-specific fine-tuning, as detailed in Appendix~\ref{app:chapter-performance}.

This significant improvement stems from HDF1's ability to reward clinically relevant differential diagnoses, an important aspect of diagnostic reasoning that flat metrics penalize. These domain-specialized models may not always identify the precise subcategory-level diagnosis, but they consistently generate a DDx list that is coherent within the correct medical domain. Our findings demonstrate that conventional benchmarks, by focusing solely on exact-match accuracy, consistently underestimate models that produce diagnostically sound and clinically useful suggestions. This behavioral profile may arise because medical domain-tuning strengthens a model's understanding of broad clinical categories, while the long tail of rare or highly specific sub-diagnoses remains underrepresented in training data. These findings suggest that smaller, domain-focused models could potentially offer a more reliable and effective solution for healthcare institutions than larger, general-purpose models. 

The substantial ranking shifts observed above raise important questions about the underlying diagnostic patterns that distinguish these models. To address this, we leverage the hierarchical analysis capabilities of H-DDx to examine how models achieve their diagnostic performance across different levels of specificity.

\subsection{Enhanced Interpretability of Model Performance}
\label{sec:rq2}

\begin{figure}[t]
  \centering
  \includegraphics[width=1.0\linewidth]{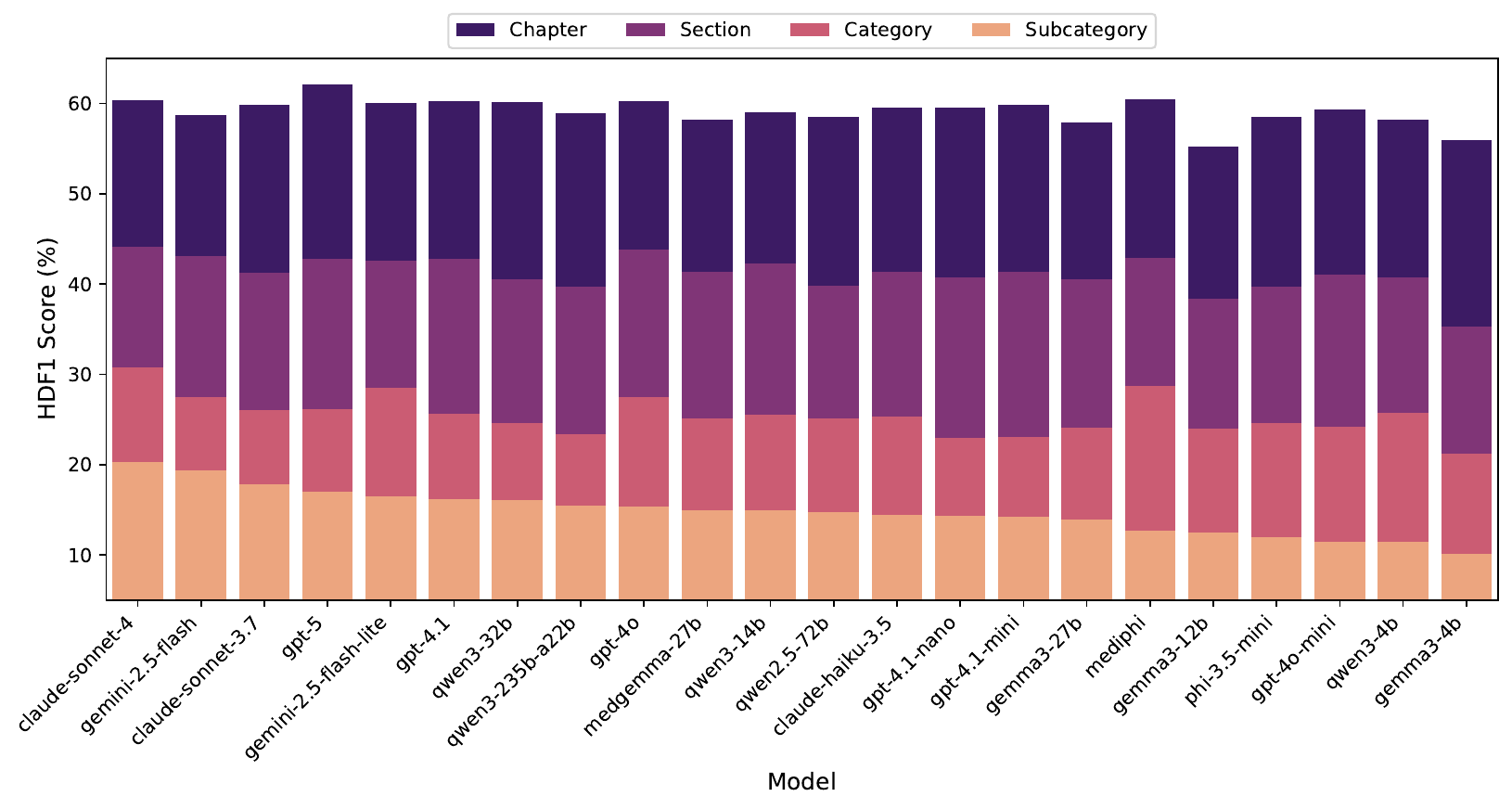}
  \caption{Hierarchical cascade pattern in diagnostic performance across ICD-10 levels. The overlapping bars show HDF1 scores calculated for each hierarchy level, with models sorted by their performance at the Subcategory level.}
  \label{fig:fig3}
\end{figure}

Beyond providing scores, H-DDx serves as an interpretive framework that reveals the diagnostic reasoning patterns of models. By analyzing performance across hierarchy levels, we identified fundamental limitations shared by all evaluated models. 
As shown in Figure~\ref{fig:fig3}, HDF1 scores are calculated independently for each ICD-10 taxonomy level (Chapter, Section, Category, and Subcategory) across all models. This visualization reveals a universal hierarchical cascade pattern where all models exhibit consistent performance degradation, with performance at the Chapter level reaching approximately 60\%, declining progressively through Section ($\sim$40\%), Category ($\sim$30\%), to Subcategory (10-20\%).

This hierarchical cascade pattern exposes fundamental limitations in current evaluation practices. This nuanced performance distribution remains invisible to conventional flat metrics. Exact-match approaches fail to recognize diagnostically relevant but inexact answers, while LLM-based judges, despite potentially recognizing partial correctness, lack consistent clinical criteria for evaluation. 
In contrast, HDF1 leverages the medically established ICD-10 taxonomy to provide systematic and reproducible partial credit, acknowledging that correctly identifying the broader medical domain represents genuine diagnostic capability even when the specific diagnosis is missed.

Our hierarchical analysis reveals distinct diagnostic profiles and explains the ranking shifts observed in Section~\ref{sec:rq1}. MediPhi, despite its low flat-metric accuracy, demonstrates strong performance at Chapter, Section, and Category levels, confirming that domain-specific fine-tuning successfully enhances broad clinical reasoning even when rare or highly specific diagnoses remain challenging. Comparing general-purpose models, GPT-4o exhibits a smooth performance degradation across all levels, while GPT-5 shows irregular performance with strong results at Chapter and Subcategory but weaker results at intermediate levels. This pattern suggests that consistent hierarchical reasoning may be more valuable than sporadic accuracy at specific levels.

By decomposing model performance across diagnostic hierarchies, H-DDx transforms evaluation from a binary assessment into a comprehensive analysis of clinical reasoning capabilities. 
Consequently, H-DDx enables more informed model selection by matching specific clinical requirements with model capabilities, whether for initial diagnostic screening or precise subspecialty identification.
Although these aggregate patterns reveal important insights, individual case studies best illustrate the clinical value of H-DDx.

\subsection{Clinical Validation of HDF1 through Case Studies}
\label{sec:rq3}

To demonstrate the clinical significance of our HDF1 score, we present two distinct case studies,
with additional cases provided in Appendix \ref{app:additional_case}. These cases exemplify how conventional flat metrics, such as Top-5 Accuracy, can be misleading or uninformative, while HDF1 accurately assesses a model's clinical reasoning. The first case highlights a scenario where a domain-specialized model demonstrates superior clinical reasoning over a general-purpose proprietary model, despite having a lower accuracy score. The second case illustrates an instance where HDF1 reveals the impact of medical fine-tuning by distinguishing between a base model and its specialized counterpart.

\begin{table}[t]
\centering
\caption{Comparative analysis of model performance on two distinct clinical scenarios. HDF1 reveals nuanced clinical reasoning capabilities that Top-5 Accuracy misses. Higher HDF1 scores are shown in bold.}
\label{tab:table3}
\resizebox{\textwidth}{!}{%
  \begin{tabular}{@{}lllcccl@{}}
\toprule
\textbf{Patient Case} & \textbf{Final Diagnosis} & \textbf{Ground-Truth DDx} & \textbf{Model} & \textbf{Top-5 Acc.} & \textbf{HDF1} & \textbf{Predicted DDx Set \& ICD-10 Codes} \\
\midrule
\multirow{2}{*}{\makecell[l]{\textbf{Case 1} \\\\ 48/F with complex \\ respiratory history}} & \multirow{2}{*}{\makecell[l]{\\\\Bronchiectasis \\ (J47)}} & \multirow{2}{*}{\makecell[l]{\\Bronchitis (J40) \\ Pulm. neoplasm (C34) \\ Tuberculosis (A15) \\ Bronchiectasis (J47) \\ Ac. pulm. edema (J81.0)}} & GPT-4o & 1.0 & 0.2069 & \makecell[l]{Bronchiectasis (J47) \\ Pulmonary Hemorrhage (R04.89) \\ COPD (J44.9) \\ Pulmonary Embolism (I26) \\ Infectious Pneumonia (J16.8)} \\
\cmidrule(l){4-7}
& & & MediPhi & 0.0 & \textbf{0.5714} & \makecell[l]{Pneumonia (J18) \\ Bronchitis (J40) \\ Tuberculosis (A15-A19) \\ Pulmonary Embolism (I26) \\ Lung Cancer (C34.90)} \\
\midrule
\multirow{2}{*}{\makecell[l]{\textbf{Case 2} \\\\ 62/F with osteoporosis \\ \& severe chest pain}} & \multirow{2}{*}{\makecell[l]{\\\\Spontaneous \\ rib fracture \\ (S22.9)}} & \multirow{2}{*}{\makecell[l]{\\Spont. rib fracture (S22.9) \\ Bronchitis (J40) \\ Whooping cough (A37) \\ Pulm. embolism (I26) \\ NSTEMI/STEMI (I21)}} & Gemma3-27B & 0.0 & 0.1935 & \makecell[l]{Pleurisy (R09.1) \\ Musculoskeletal Chest Pain (R07.82) \\ Costochondritis (M94.0) \\ Pulmonary Embolism (I26) \\ Pericarditis (I30)} \\
\cmidrule(l){4-7}
& & & MedGemma-27B & 0.0 & \textbf{0.5000} & \makecell[l]{Pulmonary Embolism (I26) \\ Pleurisy (R09.1) \\ Costochondritis (M94.0) \\ \textbf{Rib Fracture (S22.3)} \\ Myocardial Infarction (I21)} \\
\bottomrule
\end{tabular}
}
\end{table}

\paragraph{Case Study 1: Specialist Coherence over Generalist Accuracy}

Our first case involves a 48-year-old female with a complex respiratory history, including cystic fibrosis and rheumatoid arthritis. The ground-truth diagnosis is Bronchiectasis (J47). As detailed in Table~\ref{tab:table3}, the Top-5 Accuracy metric provides a potentially misleading assessment. The general-purpose model, GPT-4o, correctly includes Bronchiectasis and achieves a perfect Top-5 Accuracy of 1.0. In contrast, the domain-specialized model, MediPhi, fails to list the exact term, resulting in an Accuracy of 0.0. Based on this metric alone, GPT-4o's performance appears far superior.

However, HDF1 offers a contrasting evaluation, assigning MediPhi a substantially higher score (0.5714) compared to GPT-4o (0.2069). A clinical review validates this assessment. The list from MediPhi, including diagnoses like Pneumonia and Bronchitis, represents a clinically coherent differential for a patient whose underlying conditions, such as cystic fibrosis, increase susceptibility to respiratory infections. HDF1 captures this clinical relevance by recognizing that these suggestions are taxonomically close to the ground-truth. For instance, Pneumonia (J18) and Bronchitis (J40) fall within the same ICD-10 chapter (J00-J99) as Bronchiectasis (J47), allowing the framework to reward the model for correctly identifying the affected organ system. Conversely, GPT-4o's DDx set, despite containing the correct answer, is less focused, including a symptom (Pulmonary Hemorrhage) rather than a distinct diagnosis. This case illustrates how HDF1 prioritizes the clinical utility of the entire DDx set over a single correct prediction.

\paragraph{Case Study 2: Revealing the Impact of Fine-Tuning in Accuracy Failures}

The second case features a 62-year-old female with osteoporosis and intense coughing fits presenting with severe chest pain. The ground-truth is Spontaneous rib fracture (S22.9). In this scenario, both the base model (Gemma3-27B) and its medically fine-tuned version (MedGemma-27B) failed to identify the correct diagnosis, resulting in a Top-5 Accuracy of 0.0 for both. Here, the flat metric fails to differentiate between their performance, suggesting they are equally poor.

However, HDF1 effectively quantifies the significant improvement gained from domain specialization, scoring MedGemma (0.5000) far higher than its base model (0.1935). A clinical review of the outputs reveals why. MedGemma generated a comprehensive differential diagnosis that included critical emergencies like Myocardial Infarction and Pulmonary Embolism, alongside the highly relevant diagnosis of Rib Fracture. While a flat metric would miss this, HDF1's hierarchical approach recognizes that Rib Fracture is taxonomically adjacent to the ground-truth, Spontaneous rib fracture (S22.9), as both belong to the same ICD-10 category (S22). In stark contrast, the base Gemma3 model failed to identify the correct diagnosis and missed a key emergency. This case effectively demonstrates HDF1's ability to measure the qualitative leap in clinical reasoning from fine-tuning and to reward clinically correct diagnoses that simpler metrics overlook.

\section{Conclusion}
In this paper, we present \textbf{H-DDx}, a novel hierarchical evaluation framework designed to address the limitations of conventional flat metrics in assessing LLMs for DDx. Existing metrics, such as Top-k accuracy, fail to distinguish between clinically relevant near-misses and diagnostically distant errors, thereby providing an incomplete picture of a model's true utility. H-DDx overcomes this by mapping free-text diagnoses to the ICD-10 taxonomy and applying the HDF1 that credits predictions hierarchically close to the ground-truth. Through a large-scale evaluation of 22 leading LLMs, we demonstrated that conventional flat metrics systematically underestimate the performance of clinically coherent models, particularly domain-specialized ones. Furthermore, H-DDx enhances interpretability by enabling a hierarchical analysis of error patterns, revealing that LLMs often correctly identify the broader clinical context even when the precise diagnosis is missed.

\section*{Limitation and Future Work} 
Our study has several limitations. First, our evaluation uses the DDXPlus synthetic dataset. While it is the largest public benchmark with over 1.3 million cases, validation on real-world clinical data remains essential. Second, the H-DDx framework relies on ICD-10, which, despite being the global standard ensuring reproducibility, may not capture all clinical nuances. Third, we evaluate static differential diagnosis lists rather than the sequential reasoning inherent to clinical practice. Nevertheless, H-DDx demonstrates significant improvements over flat metrics by recognizing clinically relevant near-misses and revealing domain-specialized models' strengths. Our mapping pipeline's high accuracy (93.1\%) further validates the approach, establishing a strong baseline that can be iteratively refined with expert clinical review.

Future work will pursue three directions. First, we will validate H-DDx on real-world EHR data to establish correlations between HDF1 scores and clinical utility. Second, we will explore richer ontologies like SNOMED CT and data-driven hierarchies to capture additional clinical nuances. Third, we will extend the framework to evaluate interactive diagnostic agents that iteratively refine diagnoses through information gathering. These advances will support the development of AI systems that meaningfully augment clinical decision-making.

\printbibliography

\medskip

\small

\newpage
\appendix

\section{Mapping Pipeline Details}
\label{app:mapping-pipeline}
This section provides a detailed, step-by-step description of our pipeline for mapping free-text diagnoses to standardized ICD-10 codes. The pipeline is designed for high accuracy by combining efficient embedding-based retrieval with the nuanced understanding of a large language model for reranking.

\subsection{Candidate Retrieval}
The initial stage focuses on retrieving a set of relevant ICD-10 code candidates for a given free-text diagnosis.

\paragraph{Knowledge Base Construction} We first constructed a knowledge base from the official ICD-10 hierarchical data. This process involved parsing the entire taxonomy, including chapters, sections, and individual diagnostic codes. To maximize coverage, we extracted not only the primary descriptive name for each code but also all associated inclusion terms (i.e., synonyms and related conditions). This resulted in a comprehensive set of \{ICD Name, ICD Code\} pairs that serves as the target for our mapping.

\begin{table}[h]
\centering
\caption{Embedding model performance}
\resizebox{0.7\linewidth}{!}{

\begin{tabular}{lS[table-format=1.4]S[table-format=1.4]S[table-format=1.4]}
\toprule
\textbf{Embedding Model} & {\textbf{Top-1 Acc}} & {\textbf{Top-5 Acc}} & {\textbf{Top-15 Acc}} \\
\midrule
biobert-v1.1 & 0.4653 & 0.6931 & 0.7525 \\
BioSimCSE-BioLinkBERT-BASE & 0.5446 & 0.7030 & 0.8218 \\
Qwen3-Embedding-0.6B & 0.5842 & 0.8020 & 0.9307 \\
pubmedbert-base-embeddings & 0.6436 & 0.8218 & 0.9109 \\
text-embedding-3-large & \textbf{0.7129} & \textbf{0.9208} & \textbf{0.9901} \\
\bottomrule
\end{tabular}
}
\label{tab:table4}
\end{table}

\paragraph{Embedding and Indexing} To select the best embedding model for our retrieval task, we benchmarked a diverse set of models. This included models pre-trained on biomedical corpora, such as \texttt{biobert-v1.1}, \texttt{BioSimCSE-BioLinkBERT-BASE}, and \texttt{pubmedbert-base-embeddings}, as well as general-purpose models like \texttt{Qwen3-Embedding-0.6B} and OpenAI's \texttt{text-embedding-3-large}. As shown in Table~\ref{tab:table4}, \texttt{text-embedding-3-large} significantly outperformed the others, achieving the highest Top-1 retrieval accuracy. Consequently, each unique ICD name in our knowledge base was embedded using this model. These high-dimensional vector representations were pre-computed and stored as a NumPy array, creating an efficient index for fast similarity searches. For each free-text diagnosis, we embed it using the same model and perform a cosine similarity search against this index to retrieve the top 15 candidates.

\subsection{LLM-based Reranking}
The second stage uses a powerful LLM to select the single best match from the candidate list generated in the previous stage.

\paragraph{Model and Prompting} We use \texttt{gpt-4o} as our reranker. The model is provided with a carefully structured prompt containing the original free-text diagnosis and the list of the 15 candidate ICD names retrieved in the first stage. The prompt, detailed below, instructs the model to act as a deterministic assistant, select exactly one item from the provided candidate list, and return its choice in a structured JSON format. This strict output formatting is crucial for ensuring the reliability and programmatic usability of the model's response.

\begin{promptbox}[verbatim]{System Prompt}
You are a careful, deterministic assistant. \\
Given one pathology/disease name and a list of candidate disease names, \\
you must select exactly one item strictly from the given \\
candidate list. Return JSON with a single key icd\_name. \\
Do not add any other keys.
\end{promptbox}

\begin{promptbox}[verbatim]{User Prompt Example}
Patient pathology: Chronic bronchitis \\

Candidates (choose exactly one from these names): \\
1. Chronic tracheobronchitis \\
2. Chronic emphysematous bronchitis \\
3. chronic obstructive bronchitis \\
4. Unspecified chronic bronchitis \\
5. Simple chronic bronchitis \\
6. Chronic asthmatic (obstructive) bronchitis \\
7. Chronic bronchitis NOS \\
8. Acute bronchitis \\
9. chronic emphysematous bronchitis \\
10. Mucopurulent chronic bronchitis \\
11. Chronic cough \\
12. chronic bronchitis with airway obstruction \\
13. chronic bronchitis with emphysema \\
14. Mixed simple and mucopurulent chronic bronchitis \\
15. chronic asthmatic (obstructive) bronchitis \\

Instruction: Pick the single best candidate by name only.  \\
Output JSON only. \\
\end{promptbox}

\paragraph{Final Selection} The JSON output from the LLM, which contains the chosen ICD name, is then parsed. We match this name back to the candidate list to identify the corresponding ICD-10 code. This code is considered the final, standardized mapping for the input free-text diagnosis. This multi-step process ensures that the final mapping is not only semantically similar but also clinically appropriate as judged by a state-of-the-art LLM.

\subsection{Pipeline Validation and Error Analysis}
To validate the pipeline's accuracy, we benchmarked it against a set of 101 diagnoses that were manually mapped to ICD-10 codes by clinicians. The retrieval-only stage achieved a Top-1 accuracy of 71.3\%. By adding the LLM reranking stage, the accuracy significantly increased to 93.1\%.

A qualitative analysis of the remaining errors revealed that they were not random but consisted of systematic, clinically adjacent near-misses. Table~\ref{tab:table5} shows incorrect Top-1 predictions from our validation set.

\begin{table}[h]
\centering
\caption{Error Analysis: Incorrect Top-1 Predictions.}
\resizebox{\linewidth}{!}{

\begin{tabular}{|l|c|l|c|}
\hline
\textbf{Disease name} & \textbf{ICD-10 code (by clinicians)} & \textbf{Mapped Disease name} & \textbf{Mapped ICD-10 code} \\
\hline
Alcoholic steatohepatitis & K70.0 & Alcoholic hepatitis & K70.1 \\
Cervical disc herniation & M50.0 & Other cervical disc displacement & M50.2 \\
Drug-induced acute kidney injury & N17.0 & Drug- and heavy-metal-induced tubulo-interstitial and tubular conditions & N14 \\
Lumbar compression fracture & S32.0 & Wedge compression fracture of unspecified lumbar vertebra & S32.000 \\
sleep disorder & G47.9 & Sleep disorders & G47 \\
Drug-induced acute kidney injury & N17.0 & Drug- and heavy-metal-induced tubulo-interstitial and tubular conditions & N14 \\
Transient fatigue & R53.83 & Heat fatigue, transient & T67.6 \\
\hline
\end{tabular}%

}
\label{tab:table5}
\end{table}

In the majority of observed failure cases, the predicted code was either clinically related or correct at the immediate parent level in the ICD-10 taxonomy. These minor taxonomic shifts do not compromise the integrity of our hierarchical evaluation, confirming the pipeline's robustness for its intended purpose.

\section{Evaluation Details}
\label{app:reproducibility}
To ensure the reproducibility of our results, this section provides comprehensive details of our evaluation setup. This includes the prompts used for querying the models, API parameters, and any data preprocessing steps.

\subsection{Models}
The following is a comprehensive list of the 22 LLMs evaluated in this study, including their specific versions. We evaluated 11 proprietary models and 11 open-source models. The model checkpoints for open-source models are downloaded from \url{https://huggingface.co/}. The specific models and their sources are as follows:

\paragraph{Proprietary Models}
\begin{itemize}
    \item \textbf{Claude-Haiku-3.5}: \texttt{claude-3-5-haiku-20241022}
    \item \textbf{Claude-Sonnet-3.7}: \texttt{claude-3-7-sonnet-20250219}
    \item \textbf{Claude-Sonnet-4}: \texttt{claude-sonnet-4-20250514}
    \item \textbf{Gemini-2.5-Flash-Lite}: \texttt{gemini-2.5-flash-lite}
    \item \textbf{Gemini-2.5-Flash}: \texttt{gemini-2.5-flash}
    \item \textbf{GPT-4o-mini}: \texttt{gpt-4o-mini-2024-07-18}
    \item \textbf{GPT-4o}: \texttt{gpt-4o-2024-08-06}
    \item \textbf{GPT-4.1-nano}: \texttt{gpt-4.1-nano-2025-04-14}
    \item \textbf{GPT-4.1-mini}: \texttt{gpt-4.1-mini-2025-04-14}
    \item \textbf{GPT-4.1}: \texttt{gpt-4.1-2025-04-14}
    \item \textbf{GPT-5}: \texttt{gpt-5-chat-latest}
\end{itemize}

\paragraph{Open-source Models}
\begin{itemize}
    \item \textbf{Gemma3-4B}: \url{https://huggingface.co/google/gemma-3-4b-it}
    \item \textbf{Gemma3-12B}: \url{https://huggingface.co/google/gemma-3-12b-it}
    \item \textbf{Gemma3-27B}: \url{https://huggingface.co/ISTA-DASLab/gemma-3-27b-it-GPTQ-4b-128g}
    \item \textbf{Qwen2.5-72B}: \url{https://huggingface.co/Qwen/Qwen2.5-72B-Instruct-AWQ}
    \item \textbf{Qwen3-4B}: \url{https://huggingface.co/Qwen/Qwen3-4B}
    \item \textbf{Qwen3-14B}: \url{https://huggingface.co/Qwen/Qwen3-14B}
    \item \textbf{Qwen3-32B}: \url{https://huggingface.co/Qwen/Qwen3-32B-AWQ}
    \item \textbf{Qwen3-235B-A22B}: \\ \url{https://huggingface.co/QuantTrio/Qwen3-235B-A22B-Instruct-2507-AWQ}
    \item \textbf{Phi-3.5-Mini}: \url{https://huggingface.co/microsoft/Phi-3.5-mini-instruct}
\end{itemize}

\paragraph{Medical Fine-tuned Models}
\begin{itemize}
    \item \textbf{MedGemma-27B}: \url{https://huggingface.co/google/medgemma-27b-text-it}
    \item \textbf{MediPhi}: \url{https://huggingface.co/microsoft/MediPhi-Instruct}
\end{itemize}

\subsection{Prompts for DDx Generation}
All models were evaluated using a consistent zero-shot prompting strategy with structured input and output formats.

\begin{promptbox}[verbatim]{System Prompt}
You are a diagnostic assistant. \\
Based on the patient's clinical information, provide a differential diagnosis. \\
Return the response in JSON format.
\end{promptbox}

The user prompt instructed models to provide the top 5 most likely differential diagnoses based on patient information (sex, age, and clinical evidence in question-answer format):
\begin{promptbox}[verbatim]{User Prompt}
Based on the following patient information, provide the top 5 most likely differential diagnoses in probability order. \\
For each diagnosis, provide the English medical term. Do not include any other text in your response without the JSON format. \\

Patient Information: \\
\{patient\_information\} \\

Please provide the response in a JSON object with a single key "diagnoses", which is a list of text. \\
Example format: \\
\{\{ \\
  "diagnoses": ["Pneumonia", "Bronchitis", "Influenza", "URTI", "Asthma"] \\
\}\}
\end{promptbox}

\subsection{Inference Configuration}
To maintain consistency, we used standardized inference parameters for all models.
\begin{itemize}
    \item \textbf{Temperature:} Set to \texttt{0.1} for all models to encourage deterministic and factual responses.
    \item \textbf{Maximum Tokens:} The maximum number of generated tokens was set to \texttt{1024} for all models.
    \item \textbf{Reasoning Effort:} For Gemini models, the \texttt{reasoning\_effort} parameter was explicitly set to \texttt{"none"}.
    \item \textbf{Output Format:} All models were constrained to generate valid JSON objects. For proprietary APIs (OpenAI, Anthropic, Gemini), we utilized their native JSON output modes. For open-source models served via a vLLM-based inference server, we enforced the JSON schema using guided generation.
\end{itemize}

\section{Dataset Statistics}
\label{app:dataset-statistics}

\paragraph{ICD-10 Chapter Distribution}

Table~\ref{tab:table6} presents the distribution of ground-truth differential diagnosis across ICD-10 chapters in our test set, providing insights into the medical domain coverage of the DDXPlus subset.

\begin{table}[t]
\centering
\caption{Distribution of ground-truth DDx by ICD-10 chapter in the evaluation dataset.}
\resizebox{0.9\linewidth}{!}{
\begin{tabular}{l r}
\toprule
\textbf{ICD-10 Chapter} & \textbf{Number of Diagnosis} \\
\midrule
Diseases of the respiratory system (J00-J99) & 532 \\
Diseases of the circulatory system (I00-I99) & 454 \\
Certain infectious and parasitic diseases (A00-B99) & 248 \\
Diseases of the nervous system (G00-G99) & 189 \\
Injury, poisoning and other consequences of external causes (S00-T88) & 168 \\
Diseases of the blood and blood-forming organs (D50-D89) & 149 \\
Diseases of the digestive system (K00-K95) & 121 \\
Mental and behavioural disorders (F01-F99) & 114 \\
Neoplasms (C00-D49) & 111 \\
Diseases of the musculoskeletal system and connective tissue (M00-M99) & 27 \\
Symptoms, signs and abnormal clinical and laboratory findings (R00-R99) & 18 \\
Diseases of the ear and mastoid process (H60-H95) & 9 \\
\bottomrule
\end{tabular}

}
\label{tab:table6}
\end{table}

\section{Performance by Medical Specialty}
\label{app:chapter-performance}

This section examines the impact of medical domain fine-tuning across different ICD-10 chapters, revealing substantial heterogeneity in performance gains across clinical specialties. We analyze all chapters with at least 100 test cases, sorted by patient count in descending order.

\begin{table}[b]
\centering
\caption{Performance comparison of base and medically-tuned models by ICD-10 Chapter. The values in parentheses indicate the change in HDF1 score after medical domain tuning.}
\resizebox{\linewidth}{!}{

\begin{tabular}{lcccccccccc}
\toprule
Model & Overall & J00-J99 & I00-I99 & A00-B99 & G00-G99 & S00-T88 & D50-D89 & K00-K95 & F01-F99 & C00-D49 \\
&         & (n=532) & (n=454) & (n=248) & (n=189) & (n=168) & (n=149) & (n=121) & (n=114) & (n=111) \\
\midrule
Gemma-3-27B & 32.25 & 33.31 & 30.44 & 10.68 & 23.75 & 19.50 & 10.03 & 31.36 & 18.90 & 33.93 \\
MedGemma-27B & 33.10 & 30.81 & 31.13 & 10.66 & 20.94 & 20.56 & 13.58 & 27.21 & 21.16 & 33.47 \\
& (\textcolor{blue}{+0.85}) & (\textcolor{red}{-2.50}) & (\textcolor{blue}{+0.69}) & (\textcolor{red}{-0.02}) & (\textcolor{red}{-2.81}) & (\textcolor{blue}{+1.06}) & (\textcolor{blue}{+3.55}) & (\textcolor{red}{-4.15}) & (\textcolor{blue}{+2.26}) & (\textcolor{red}{-0.46}) \\
\midrule
Phi-3.5-mini & 31.87 & 37.72 & 24.33 & 11.45 & 13.86 & 17.18 & 10.82 & 21.58 & 16.24 & 28.64 \\
MediPhi & 35.26 & 43.38 & 25.17 & 8.01 & 13.09 & 15.90 & 20.97 & 29.65 & 18.76 & 30.55 \\
& (\textcolor{blue}{+3.39}) & (\textcolor{blue}{+5.66}) & (\textcolor{blue}{+0.84}) & (\textcolor{red}{-3.44}) & (\textcolor{red}{-0.77}) & (\textcolor{red}{-1.28}) & (\textcolor{blue}{+10.15}) & (\textcolor{blue}{+8.07}) & (\textcolor{blue}{+2.52}) & (\textcolor{blue}{+1.91}) \\
\bottomrule
\end{tabular}
}
\label{tab:table7}
\end{table}

Table~\ref{tab:table7} presents the performance comparison across different ICD-10 chapters. The results demonstrate that medical fine-tuning produces highly variable effects across clinical specialties. MediPhi shows substantial overall improvement (+3.39 HDF1) over its base model, with particularly strong gains in diseases of the blood and blood-forming organs (D50-D89: +10.15), digestive system (K00-K95: +8.07), and respiratory system (J00-J99: +5.66). However, the same fine-tuning process leads to performance degradation in certain infectious and parasitic diseases (A00-B99: --3.44), highlighting that domain specialization can be detrimental for specific medical areas. MedGemma-27B exhibits more modest and mixed results, with a marginal overall gain (+0.85 HDF1). While it improves on blood disorders (D50-D89: +3.55), it shows notable performance drops in key areas, including the digestive system (K00-K95: --4.15), nervous system (G00-G99: --2.81), and respiratory system (J00-J99: --2.50) chapters.

These findings underscore that the effectiveness of medical domain fine-tuning is not universal but rather depends critically on the alignment between the fine-tuning data distribution and the specific knowledge requirements of each medical specialty.

\section{Model Ranking Analysis}
\label{app:ranking_shift}
\begin{figure}[t]
\centering
\includegraphics[width=0.7\linewidth]{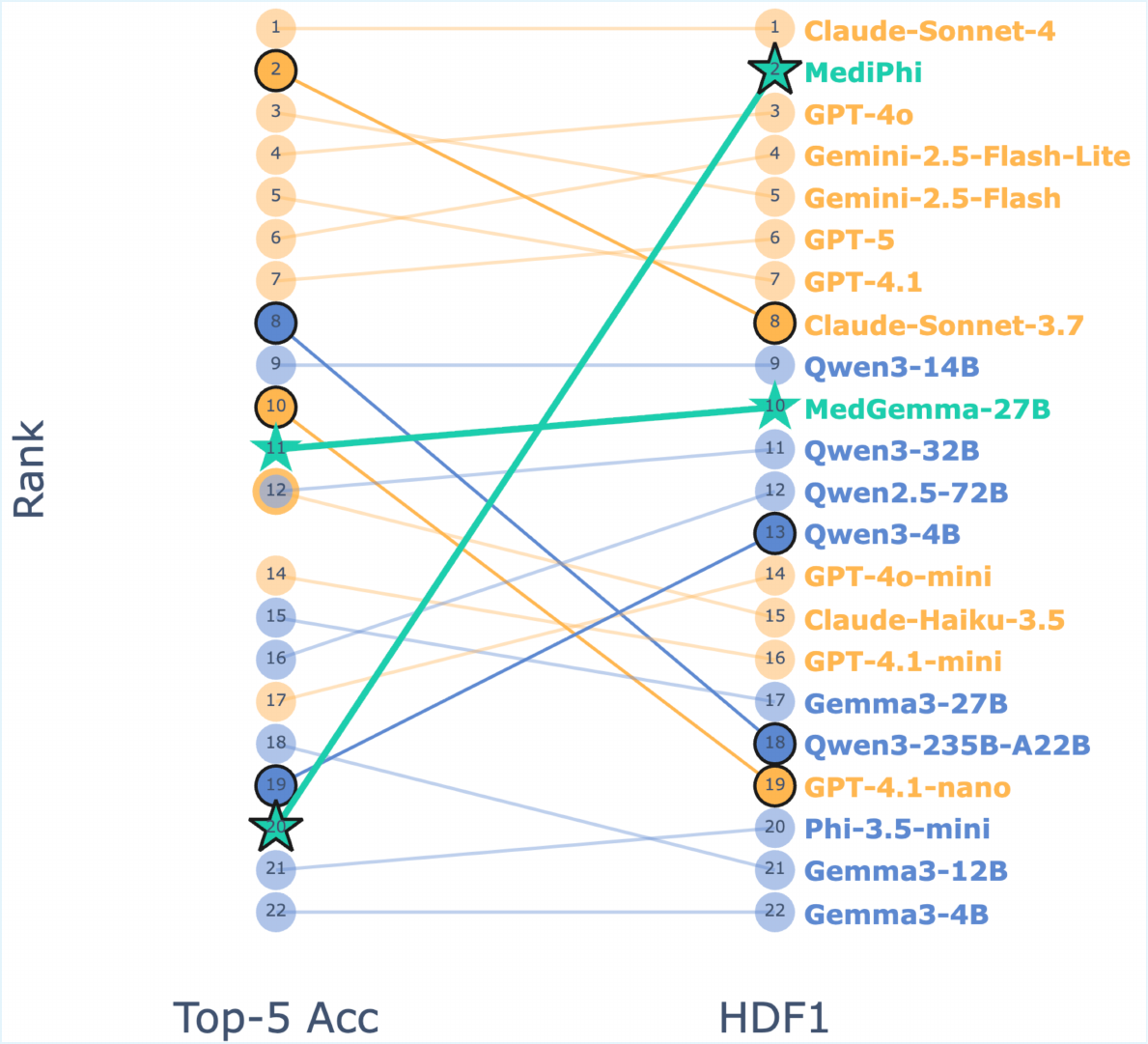}
\caption{Shift in model rankings from flat (Top-5 Accuracy) to hierarchical (HDF1) evaluation. The figure illustrates the rank changes when moving from a conventional accuracy metric (left) to the proposed hierarchical score (right). Models are color-coded as proprietary (yellow), open-source (blue), and medically fine-tuned (green). Note the significant rank improvement of medically fine-tuned models like MediPhi, which highlights HDF1's ability to better capture clinical relevance.}
\label{fig:fig4}
\end{figure}

Figure~\ref{fig:fig4} illustrates how our hierarchical evaluation framework substantially reorders model rankings compared to conventional flat metrics. The transition from Top-5 Accuracy to HDF1 reveals that domain-specialized models like MediPhi achieve marked ranking improvements. This shift demonstrates HDF1's ability to capture clinically meaningful performance characteristics, such as rewarding coherent but inexact diagnoses, a nuance that flat metrics consistently overlook.

\section{Additional Case Studies on Hierarchical Evaluation}
\label{app:additional_case}

This section presents two additional case studies that further demonstrate the nuanced evaluation capabilities of the H-DDx framework, as shown in Table~\ref{tab:table8}. These cases reveal how HDF1 can differentiate clinical reasoning quality even when flat metrics show identical results.

\paragraph{Case Study 3: Distinguishing Clinical Coherence Despite Equal Accuracy} Our third case involves a 21-year-old female presenting with upper respiratory symptoms. The patient lives in crowded conditions, works at a daycare, and presents with fever, sore throat, productive cough, nasal congestion, and diffuse muscle pain. The ground-truth diagnosis is Upper Respiratory Tract Infection (URTI, J06.9).

In Case 3, both models achieve a perfect Top-5 Accuracy of 1.0. Notably, while Gemma3-4B did not explicitly predict ``URTI,'' the LLM judge deemed ``Acute Pharyngitis'' (J02) as semantically equivalent to URTI, highlighting the ambiguity inherent in LLM-based matching. This exemplifies a key limitation of flat metrics that rely on subjective semantic matching. Claude-Sonnet-4 directly identifies URTI and generates a coherent differential entirely within the respiratory domain, including Influenza, Sinusitis, Pneumonia, and Bronchitis, all of which are plausible given the patient's presentation.

In contrast, despite receiving the same Top-5 Accuracy score through the judge's lenient interpretation, Gemma3-4B's differential includes Viral Meningitis (A87) and Cellulitis (L03.90), which are clinically inconsistent with the primary respiratory presentation. HDF1 reveals this dramatic difference in clinical reasoning quality (0.1935 vs 0.7692), demonstrating that accuracy alone, especially when determined through ambiguous semantic matching, fails to capture the clinical utility of the entire differential diagnosis set.

\paragraph{Case Study 4: Gradations in Clinical Failure} 

The fourth case features a 46-year-old male with joint pain, mouth ulcers, shortness of breath, and fatigue. The ground-truth diagnosis is Systemic Lupus Erythematosus (SLE, M34). Both models fail to identify SLE, resulting in a Top-5 Accuracy of 0.0 for both.

While both models fail completely by flat metric standards, HDF1 reveals meaningful differences in their clinical reasoning (0.1212 vs 0.1714). GPT-4o-mini demonstrates marginally better clinical judgment by including systemic rheumatological conditions such as Rheumatoid Arthritis (M06.9) and Polymyalgia Rheumatica (M35.3), which share the autoimmune nature and systemic involvement characteristic of SLE. Both models include musculoskeletal conditions (Fibromyalgia, Ankylosing Spondylitis), but GPT-4o-mini's inclusion of Chronic Fatigue Syndrome acknowledges the systemic nature of the presentation. In contrast, Gemma3-12B includes less relevant diagnoses such as Hypertension and TMJ Disorder. While neither model performs well, HDF1 captures these subtle differences in clinical reasoning that would be invisible to binary accuracy metrics.

\paragraph{Key Insights from Additional Cases} These additional case studies reinforce two critical findings:

\begin{enumerate}
    \item \textbf{Equal accuracy does not imply equal clinical utility.} Case 3 shows that models with identical Top-5 Accuracy can have vastly different clinical reasoning quality, with HDF1 scores differing by nearly 4-fold.
    
    \item \textbf{Hierarchical evaluation provides granularity even in failure.} Case 4 demonstrates that even when all models fail to identify the correct diagnosis, HDF1 can still distinguish degrees of clinical relevance in their differential diagnoses.
\end{enumerate}

These cases further validate that H-DDx provides a more nuanced and clinically meaningful evaluation framework compared to conventional flat metrics, enabling better assessment of LLMs for medical diagnostic tasks.

\begin{table}[t]
\centering
\caption{Additional comparative analysis demonstrating HDF1's ability to differentiate clinical reasoning quality. Higher HDF1 scores are shown in bold.}
\begin{adjustbox}{width=\textwidth,center}
  \begin{tabular}{@{}lllcccl@{}}
\toprule
\textbf{Patient Case} & \textbf{Final Diagnosis} & \textbf{Ground-Truth DDx} & \textbf{Model} & \textbf{Top-5 Acc.} & \textbf{HDF1} & \textbf{Predicted DDx Set \& ICD-10 Codes} \\
\midrule
\multirow{2}{*}{\makecell[l]{\textbf{Case 3} \\\\ 21/F with URI \\ symptoms}} & \multirow{2}{*}{\makecell[l]{\\\\URTI \\ (J06.9)}} & \multirow{2}{*}{\makecell[l]{\\URTI (J06.9) \\ Influenza (J11.1) \\ Pneumonia (J18) \\ Bronchitis (J40) \\ HIV initial (B20)}} & Gemma3-4B & 1.0 & 0.1935 & \makecell[l]{Viral Meningitis (A87) \\ Sinusitis (J01.9) \\ Acute Pharyngitis (J02) \\ COVID-19 (U07.1) \\ Cellulitis (L03.90)} \\
\cmidrule(l){4-7}
& & & Claude-Sonnet-4 & 1.0 & \textbf{0.7692} & \makecell[l]{Influenza (J11.1) \\ URTI (J06.9) \\ Sinusitis (J01.9) \\ Pneumonia (J18) \\ Bronchitis (J40)} \\
\midrule
\multirow{2}{*}{\makecell[l]{\textbf{Case 4} \\\\ 46/M with joint pain \\ \& mouth ulcers}} & \multirow{2}{*}{\makecell[l]{\\\\SLE \\ (M34)}} & \multirow{2}{*}{\makecell[l]{\\NSTEMI/STEMI (I21) \\ SLE (M34) \\ Anemia (D64.9) \\ Pulm. neoplasm (C34) \\ Acute dystonia (G24.02)}} & Gemma3-12B & 0.0 & 0.1212 & \makecell[l]{Fibromyalgia (M79.7) \\ Hypertension (I10-I1A) \\ Ankylosing Spondylitis (M45) \\ Cervical Radiculopathy (M54.12) \\ TMJ Disorder (M26.6)} \\
\cmidrule(l){4-7}
& & & GPT-4o-mini & 0.0 & \textbf{0.1714} & \makecell[l]{Fibromyalgia (M79.7) \\ Chronic Fatigue Syndrome (G93.32) \\ Rheumatoid Arthritis (M06.9) \\ Polymyalgia Rheumatica (M35.3) \\ Ankylosing Spondylitis (M45)} \\
\bottomrule
\end{tabular}
\end{adjustbox}
\label{tab:table8}
\end{table}

\end{document}